\pdfoutput=1

\documentclass[11pt]{article}

\usepackage{emnlp2021}

\usepackage{times}
\usepackage{latexsym}
\usepackage{amsmath}
\usepackage{graphicx}
\usepackage{float}
\usepackage{setspace}
\usepackage[normalem]{ulem}
\useunder{\uline}{\ul}{}

\usepackage[T1]{fontenc}

\usepackage[utf8]{inputenc}

\usepackage{microtype}

\title{Document-Level Text Simplification: Dataset, Criteria and Baseline}

 \author{\Large \textbf{Renliang Sun, Hanqi Jin, Xiaojun Wan}\\ 
Wangxuan Institute of Computer Technology, Peking University\\
Center for Data Science, Peking University\\
The MOE Key Laboratory of Computational Linguistics, Peking University\\
{\tt  sunrenliangpku@gmail.com}\\
{\tt  \{jinhanqi, wanxiaojun\}@pku.edu.cn}}

\begin{document}
\maketitle
\begin{abstract}
Text simplification is a valuable technique. However, current research is limited to sentence simplification. In this paper, we define and investigate a new task of document-level text simplification, which aims to simplify a document consisting of multiple sentences. Based on Wikipedia dumps, we first construct a large-scale dataset named D-Wikipedia and perform analysis and human evaluation on it to show that the dataset is reliable. Then, we propose a new automatic evaluation metric called D-SARI that is more suitable for the document-level simplification task. Finally, we select several representative models as baseline models for this task and perform automatic evaluation and human evaluation. We analyze the results and point out the shortcomings of the baseline models.
\end{abstract}

\section{Introduction}

Text simplification is a valuable technique that deserves to be studied in depth \cite{woodsend2011learning}. One definition of text simplification is to simplify the original text to a more understandable text, while keeping the main meaning of the original text unchanged \cite{vstajner2018data, maddela2020controllable}. It can provide convenience for non-native speakers \cite{petersen2007text, glavavs2015simplifying, paetzold2016unsupervised}, non-expert readers \cite{elhadad2007mining, siddharthan2010reformulating} and children \cite{de2010text, kajiwara2013selecting}. 

\subsection{Why it is Valuable to Study Document-level Text Simplification}

Currently, researches on text simplification focus on sentence simplification, and the existing common text simplification datasets such as Wikilarge, Wikismall, and Newsela are also designed for sentence simplification. However, various complex applications in the real world often require document-level simplification rather than sentence-level simplification. Imagining that if you want to simplify an article in Time magazine for children to read, it is very inefficient to simplify the sentences separately. Besides, sentences that are obscure and have little relation to the subject should be deleted instead of simplified. Therefore, studying document-level text simplification may be more meaningful than studying sentence-level text simplification alone. Unfortunately, the research on document-level text simplification is still scarce: there is no formal definition, no suitable dataset, and evaluation criteria.

\subsection{Similarities and Differences with Text Summarization}

Other tasks that may be related to document-level text simplification are text summarization\cite{dong2018banditsum,cao2020jointly}, paraphrasing \cite{zhao2018integrating,guo2018dynamic}, and split \& rephrase \cite{narayan2017split,surya2019unsupervised}. Obviously, the paraphrasing and split \& rephrase tasks are both sentence-level tasks.
The most closely related task is text summarization, which is also a document-level task. We use an example to illustrate the difference between text summarization and our task, as shown in Table \ref{tab:example}. We can see that text summarization does not involve rewriting text with simplified versions, though both the two tasks may filter or delete some unimportant text from the original document. 


\begin{table*}[h]
\centering
\begin{tabular}{l|l}
\hline
\begin{tabular}[c]{@{}l@{}}\small Original\\ \small article\end{tabular}    & \begin{tabular}[c]{@{}l@{}}
\small Firefighters or firemen are people whose job is to put out fires and rescue people. Besides fires, \\ \small firefighters rescue people and animals from car wrecks, collapsed buildings, stuck elevators and many\\ \small other emergencies. Firefighting is a job which requires bravery, strength, quick thinking and a wide\\ \small range of skills. Firefighters are based at a building called a `` fire station '' ( also known as a `` firehouse\\ \small '' or `` fire hall '' ). When their help is needed, they drive a vehicle called a `` fire engine '' or `` fire truck ''\\ \small  to the scene responding  code 1 code 2 or code 3. These vehicles can pump water and foam to put out\\ \small fires. Fire engines also carry ladders, cutting tools and lots of different types of rescue equipment.  Most\\ \small carry first aid kits to help people who are injured or hurt.

\end{tabular} \\ \hline
\begin{tabular}[c]{@{}l@{}}\small Document-level\\ \small simplification\end{tabular} & \begin{tabular}[c]{@{}l@{}}\small \textbf{The job of} a firefighter is to put out fires and \textbf{save lives from many emergencies}. They are based at a \\ \small building called a `` fire station ''. They drive a vehicle called a `` fire engine '' or `` fire truck '' to the\\ \small scene. The vehicle carries many types of rescue equipment \textbf{to help people in danger}.\end{tabular}                                                                                                                                                                                                                                                                                                                                                                                                                                                                                                                                     \\ \hline
\begin{tabular}[c]{@{}l@{}}\small Text\\ \small summarization\end{tabular}  & \begin{tabular}[c]{@{}l@{}}\small Firefighters or firemen are people whose job is to put out fires and rescue people and animals from\\ \small many emergencies. Firefighters are based at a building called a `` fire station ''. When their help is \\ \small needed, they drive a vehicle called a `` fire engine '' or `` fire truck '' which may carry different types of\\ \small rescue equipment to help people who are injured or hurt to the scene .\end{tabular}                                                                                                                                                                                                                                                                                                                                                                                                                                            \\ \hline
\end{tabular}
\caption{Examples for document-level simplification and text summarization. It can be seen from the \textbf{Bold} part that the simplified article not only deletes complicated and unimportant sentences from the original article but rewrites the clause, merges two sentences then simplifies them, replaces difficult words, etc. These are operations that text summarization does not require.}
\label{tab:example}
\end{table*}


\subsection{Our Contributions}

In this paper, we are committed to promoting research on document-level text simplification.
In summary, the main contributions of our work include:


\begin{itemize}
  \item [(1)] 
  We define the new task of document-level text simplification and build the D-Wikipedia dataset for research\footnote{The D-Wikipedia dataset is released at \url{https://github.com/RLSNLP/Document-level-text-simplification}. }. 
  \item [(2)]
  We propose a new automatic evaluation metric called D-SARI that is more suitable for the new task.
  \item [(3)]
  We select several representative models and perform both automatic evaluation and human evaluation. The results could serve as the baselines.
\end{itemize}

\section{Related Works}

Sentence simplification aims to rewrite an original sentence into a more straightforward sentence \cite{saggion2017automatic,sulem2018semantic}. The input and output of the model are just sentences instead of articles. Based on the English Wikipedia and the Simple English Wikipedia, many researchers have built high-quality datasets such as Wikilarge \cite{zhang2017sentence}, Wikismall \cite{zhu2010monolingual}, and so on \cite{coster2011learning,kauchak2013improving}. Based on Newsela, \newcite{xu2015problems} established the Newsela dataset. The above datasets are widely used in the field of sentence simplification.

\begin{table*}[h]
\centering
\begin{tabular}{c|cccccc}
\hline
Operation & \begin{tabular}[c]{@{}c@{}}Sentence\\ joining\end{tabular} & \begin{tabular}[c]{@{}c@{}}Sentence\\ splitting\end{tabular} & \begin{tabular}[c]{@{}c@{}}Sentence\\ deletion\end{tabular} & \begin{tabular}[c]{@{}c@{}}Sentence\\ reordering\end{tabular} & \begin{tabular}[c]{@{}c@{}}Sentence\\ addition\end{tabular} & \begin{tabular}[c]{@{}c@{}}Anaphora\\ resolution\end{tabular} \\ \hline
Percentage(\%) & 96                                                          & 84                                                            & 91                                                           & 92                                                            & 92                                                           & 92                                                             \\ \hline
\end{tabular}
\caption{Estimated percentage of articles with each of the six document-level simplification operations in the D-Wikipedia dataset. Each of these document-level operations appears in most of articles in the dataset.}
\label{tabel:proportion}
\end{table*}

Most of the early simplification models were based on statistical machine translation \cite{wubben2012sentence,narayan2014hybrid}. \newcite{nisioi2017exploring} improved the machine translation model to obtain a new simplification model. \newcite{zhang2017sentence} developed a reinforcement learning model and achieved excellent results. \newcite{vuetal2018sentence} introduced a new memory-augmented neural network to enhance the results. Two new approaches were proposed by \newcite{kriz2019complexity} to solve the problem of long and complicated simplified outputs. \newcite{scarton2018learning} and \newcite{nishihara2019controllable} investigated how to simplify sentences to different difficulty levels. \newcite{dong2019editnts} studied the three explicit edit operations in sentence simplification and proposed a new model. 
\newcite{vstajner2017sentence}, \newcite{paetzold2017massalign}, and \newcite{jiang2020neural} proposed sentence alignment methods to improve sentence simplification. \newcite{sun2020helpfulness} used the preceding and following sentences to help simplify a specifically given sentence.

There are very few works related to document-level text simplification. \newcite{alva2019cross} focused on cross-sentence transformations in text simplification and analyzed them, concluding that document-level simplification cannot be achieved by merely selecting parts of the content then simplifying individual sentences. Subsequently, \newcite{zhong2020discourse} used discourse-level factors to predict whether a sentence should be deleted and achieved good results. 

Although some previous works focus more or less on document-level information, the task of document-level text simplification is still not clearly defined, and there are no available high-quality datasets and criteria to evaluate the generated articles. 

\section{Problem Formulation}
\label{section:three}

The document-level text simplification task can be defined as follows. Given an original complex article $C$, the article consists of $n$ sentences, denoted as $C = \{S_1, S_2, ... S_n\}$. Document-level simplification aims to simplify $C$ into $m$ sentences, which form the simplified article $F$, denoted as $F = \{T_1, T_2, ... T_m\}$, and $m$ may not be equal to $n$. $F$ retains the primary meaning of $C$ and is more straightforward than $C$, making it easier for people to understand.

The operations for sentence-level simplification include word reservation and deletion, synonym replacement, etc. \cite{xu2016optimizing} Based on the work of \newcite{alva2019cross}, we define six types of document-level simplification operations, namely, sentence joining, sentence splitting, sentence deletion, sentence reordering, sentence addition, and anaphora resolution. See Appendix \ref{section:appendixA} for the specific definition and example of each operation.

In our definition, document-level simplification should allow the loss of information but should not allow the loss of important information. \newcite{zhong2020discourse} pointed out that sentence deletion is a prevalent phenomenon in document simplification. We believe that information that has little relevance to the primary meaning should be removed to improve readability.

\section{The D-Wikipedia Dataset}

\subsection{Dataset Construction}

According to the definition of document-level simplification, we built a new large-scale dataset named D-Wikipedia based on the English Wikipedia and Simple English Wikipedia. We first downloaded dumps from the official website of Wikipedia and created over 170,000 article pairs\footnote{The English Wikipedia dumps are from \url{https://dumps.wikimedia.org/enwiki} and the Simple English Wikipedia are from \url{https://dumps.wikimedia.org/simplewiki}. The version we used is 2020-08-20. We also used the WikiExtractor (\url{https://github.com/attardi/wikiextractor}) to extract and clean text from the dumps.}. Considering that it is not easy to establish a one-to-one correspondence between the contents, i.e., the subheadings, we kept only the main content, which is the abstract below the headings. Meanwhile, we considered that if the article is too long, it will occupy a large amount of memory during training. Therefore, we removed those article pairs whose original article or simplified article is longer than 1,000 words. Finally, we built a dataset containing 143,546 article pairs. The D-Wikipedia dataset not only can be used for document-level simplification research but also can be further aligned to construct a sentence-level simplification dataset.

In this work, we randomly divided the dataset into 132K article pairs as the training set, 3K article pairs as the validation set, and 8K article pairs as the test set.  There is no overlap between the training set, validation set, and test set.

\subsection{Additional Newsela Test Set}
\label{sec:Newsela}

There is also a commonly-used and high-quality corpus named Newsela that might be used for document-level simplification. Each original article in the Newsela corpus corresponds to four articles of different simplification levels. We also removed the article pairs whose original article or simplified article is longer than 1000 words. Given that the number of articles in each simplification level is less than a thousand, we only use them to build four additional test sets of different simplification levels. In addition, using the Newsela corpus requires a license\footnote{\url{https://newsela.com/data}}, while the D-Wikipedia dataset will be completely open-source.

\subsection{Statistics and Comparison}

We randomly sampled 100 article pairs from the established D-Wikipedia dataset to estimate the percentage of the articles which contain each of the six document-level simplification operations (mentioned in Section \ref{section:three}). We used Amazon Mechanical Turk to invite three workers to identify the operations in the articles, and the percentage of articles with each operation is shown in Table \ref{tabel:proportion}. It can be seen that each simplification operation appears in most of the simplified articles in the dataset. In other words, most articles involve with different simplification operations. 

\begin{figure}[h]
\centering
\includegraphics[width=6cm]{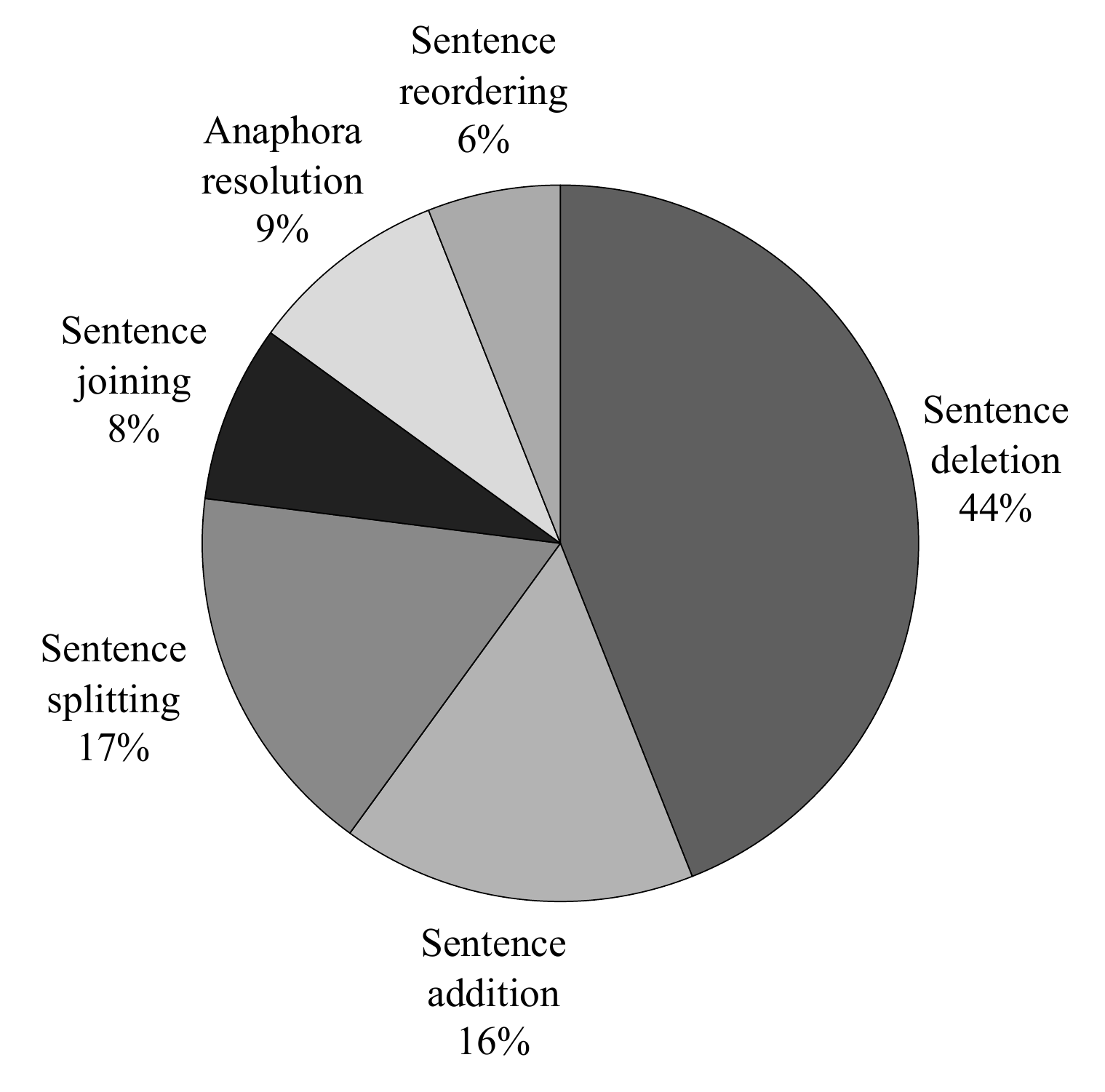}
\caption{The percentage of each simplification operation. Sentence deletion occurs most frequently, accounting for nearly half of the total simplification operations, while sentence reordering occurs least frequently, accounting for only 6\% of the total simplification operations.}
\label{pic-proportion}
\end{figure}

We also calculated the percentage of each simplification operation according to the total occurrences of the operations in the simplified articles, which is shown in Figure \ref{pic-proportion}. It can be seen that the sentence deletion operation occurs most frequently in the dataset. 

To analyze the word-level differences between the original articles and the simplified articles, following \newcite{xu2015problems}, we adopted the odds ratio method proposed by \newcite{monroe2008fightin}. The odds ratio of token $t$ between corpus $i$ and corpus $j$ is defined as:

\begin{equation}
\label{odds-ratio}
    r_{t}^{(i-j)}=\frac{y_t^i/y_t^j}{n^i/n^j}
\end{equation}

In Equation \ref{odds-ratio}, $y_t^i$ represents the count of token $t$ in corpus $i$ and $y_t^j$ represents the count of token $t$ in corpus $j$. $n^i$ and $n^j$ represent the size of corpus $i$ and corpus $j$, respectively. We have found some complex words that occur frequently as examples to show that they are sufficiently simplified, as shown in Table \ref{tab:odds-ratio}.

\begin{table}[h]
\centering
\resizebox{7.5cm}{!}{
\begin{tabular}{l|cccc}
\hline
             & \multicolumn{1}{l}{$R_{original}$} & \multicolumn{1}{l}{$R_{simple}$} & \multicolumn{1}{l}{odds ratio$\downarrow$} & \multicolumn{1}{l}{$p$-value} \\ \hline
population   & 47                            & 123                         & 0.49  & 0                         \\
including    & 68                            & 173                         & 0.49  & 0                         \\
located      & 79                            & 263                         & 0.38  & 0                         \\
metropolitan & 281                           & 904                         & 0.32  & 0                        \\ \hline
\end{tabular}}
\caption{$R_{original}$ and $R_{simple}$ indicate the ranking of the number of occurrences of the word in the original article and the simplified article, respectively. A smaller odds ratio means a greater reduction of the complex word. The closer the $p$-value of the test is to zero, the more significant the difference between the odds ratio and one.}
\label{tab:odds-ratio}
\end{table}

It can be seen from $R_{original}$ and $R_{simple}$ that the relative frequency of complex words appearing in simplified texts is much lower than in the original texts. We used a chi-square test to show if the odds ratio is significantly different from one\footnote{We use the script in \url{https://github.com/scipy/scipy/blob/v1.7.1/scipy/stats/contingency.py}.}. An odds ratio significantly lower than one means that the complex words are well simplified. The reduction of the word ``including'' may means that clauses are deleted or split into multiple sentences.

Sentence splitting is a common operation in document-level simplification. When splitting the conjoined clauses, to preserve the rhetorical relation, \newcite{siddharthan2003preserving} introduced the cue words. We calculated the odds ratio of the conjunction and the cue word, and the results are shown in Table \ref{tab:cue-word}.
We did not calculate all words because the number of occurrences of some words, such as ``hence'', was too low to be statistically meaningful.

\begin{table}[h]
\centering
\resizebox{7.5cm}{!}{
\begin{tabular}{lcccc}
\hline
\multicolumn{1}{l|}{conjunction} & although             & though               & since                & as                   \\ \hline
\multicolumn{1}{l|}{odds ratio$\downarrow$}  & 0.41                 & 0.68                 & 0.74                 & 0.64 \\
\multicolumn{1}{l|}{$p$-value}  & 0                 & 0                 & 0                 & 0
\\ \hline
\multicolumn{1}{l|}{conjunction} & and                  & or                   & but                  & \multicolumn{1}{l}{} \\ \hline
\multicolumn{1}{l|}{odds ratio$\downarrow$}  & 0.78                 & 0.93                 & 1.02                 &  \\
\multicolumn{1}{l|}{$p$-value}  &  0                &  0                &  0.04                &  \\ \hline
                                 & \multicolumn{1}{l}{} & \multicolumn{1}{l}{} & \multicolumn{1}{l}{} & \multicolumn{1}{l}{} \\ \hline
\multicolumn{1}{l|}{cue word}    & still                & then                 & also                 & however              \\ \hline
\multicolumn{1}{l|}{odds ratio$\uparrow$}  & 1.23                 & 1.18                 & 1.12                 & 0.76  \\
\multicolumn{1}{l|}{$p$-value}  &  0                &  0                &  0                &  0             \\ \hline
\end{tabular}}
\caption{The odds ratio of most conjunctions is significantly less than one, and the odds ratio of most cue words is significantly greater than one, indicating that the simplified article may contain more split sentences and the long sentences in the original article have been simplified.}
\label{tab:cue-word}
\end{table}

\begin{table*}[h]
\centering
\resizebox{14cm}{!}{
\begin{tabular}{|l|cc|cccc|}
\hline
                          & \multicolumn{2}{c|}{D-Wikipedia}           & \multicolumn{4}{c|}{Newsela corpus}                                                                                                       \\ \cline{2-7} 
                          & \multicolumn{1}{c|}{Original} & Simple     & \multicolumn{1}{c|}{Simp-1} & \multicolumn{1}{c|}{Simp-2} & \multicolumn{1}{c|}{Simp-3} & Simp-4  \\ \hline
Total articles            & \multicolumn{2}{c|}{143,546}                & 712  & 775 & 813 & 797                                                                                              \\
Total sentences           & 707,470                       & 581,513                           & 27,254                      & 34,814                      & 39,489                      & 37,329  \\
Total words               & 20,349,706                    & 11,286,155                     & 575,077                   & 613,174                   & 570,164                     & 442,173 \\ \hline
Avg words per article         & 141.76                        & 78.62                          & 807.69                      & 791.19                      & 701.31                      & 554.80   \\
-Compression ratio &                               & 0.55       & 1.05                        & 1.01                        & 0.89                        & 0.70    \\
Avg words per sent        & 28.76                         & 19.41      & 21.10                       & 17.61                        & 14.44                       & 11.85   \\
-Compression ratio &                               & 0.67       & 0.86                        & 0.72                        & 0.59                        & 0.48    \\ \hline
\end{tabular}}
\caption{Basic statistics of the D-Wikipedia dataset vs. the Newsela Simplification corpus. For the Newsela corpus, the results are different from those reported by \newcite{xu2015problems} because we deleted too long articles.}
\label{table:statistic}
\end{table*}

The D-Wikipedia dataset was also analyzed and compared with the Newsela corpus, and the results are shown in Table \ref{table:statistic}. In terms of the average number of words per article, the compression ratio of the D-Wikipedia dataset is lower than that of the Newsela corpus of any simplification level. In terms of the average number of words per sentence, the compression ratio of the D-Wikipedia dataset is between the Simp-2 level and the Simp-3 level of the Newsela corpus. 


\subsection{Human Evaluation}
\label{sec:dataset_human}

In this section, we employ human judges to evaluate the quality of the D-Wikipedia dataset. Before evaluation, we need to analyze whether the human evaluation indicators used for sentence-level simplification are suitable to evaluate document-level simplification.

In human evaluation, sentence simplification is usually evaluated from the three perspectives of simplicity, meaning, and grammar. Simplicity is the most important indicator. Simplicity indicates if the simplified sentence is simpler than the original sentence. However, we believe that this measure is not a good indicator for scoring document-level simplification. For example, if only the first sentence of the original article is simplified and the other sentences are deleted, the simplified article will be very short and simple and will get a high simplicity score. But such an article does not retain the main information of the original article, which is not what we want.
 
 
 Therefore, we propose a new indicator named O-simplicity (Overall simplicity with quality guarantee). O-simplicity indicates if the simplified article is simpler than the original article, under the condition of quality guarantee, i.e., it also should read smoothly and can retain the main meaning of the original article. As an indicator to evaluate how good the simplification is, O-simplicity is a more meaningful and comprehensive measure than the original simplicity indicator or simply averaging the simplicity, meaning, and grammar scores. 
 
 Following \newcite{sulem2018simple}, we also use the fine-grained simplicity-phrase and simplicity-structure, which measures the simplification of words and the simplification of sentence structure, respectively. In this way,  O-simplicity is an overall indicator that needs comprehensive consideration, and the other four indicators are focusing on specific aspects. More examples and scoring guidelines are given and analyzed in Appendix \ref{section:appendixC}. 
 
 
 We invited three workers to evaluate the quality of the D-Wikipedia dataset with the above measures. We randomly selected 100 article pairs from the dataset, and the five-point Likert scale is used for rating. For the results, the average O-simplicity score is 3.94, indicating the simplification of the articles is generally good. 
 The Simplicity-phrase and Simplicity-structure scores reach 4.28 and 4.23, respectively, implying that the simplified article has made considerable lexical and sentence structure simplifications compared to the original article. The grammar score achieves 4.65, probably because the simplified articles are written by humans and are easy to read. The meaning score is 3.69, indicating that the simplified article can preserve the meaning of the original article.
 



\section{The D-SARI Metric}
\label{sec:criteria}

Currently, the most commonly used automatic evaluation metric for sentence-level simplification is the SARI metric \cite{xu2016optimizing}. 
The correlation of SARI with human judgments of the simplicity indicator proved to be high in sentence-level simplification \cite{sulem2018bleu}.
However, this metric has shortcomings when used directly to evaluate document-level simplification. For better understanding, in this section, we conduct a qualitative analysis and give the following example:

\begin{table*}[h]
\centering
\begin{tabular}{|c|cccc|cccc|}
\hline
                     & SARI  & $F_{keep}$  & $P_{del}$   & $F_{add}$   & D-SARI & $D_{keep}$  & $D_{del}$   & $D_{add}$   \\ \hline
Simplified article 1 & 54.24 & 23.74 & 88.18 & 50.80 & 42.80  & 23.74 & 88.18 & 16.49 \\
Simplified article 2 & 64.90 & 33.68 & 98.08 & 62.95 & 41.00  & 11.86 & 48.19 & 62.95 \\
Simplified article 3 & 66.80 & 67.63 & 96.44 & 36.33 & 42.91  & 25.68 & 66.72 & 36.33 \\
Simplified article 4 & 49.93 & 51.39 & 91.25 & 7.14  & 48.69  & 50.06 & 88.88 & 7.14  \\ \hline
\end{tabular}
\caption{The SARI and D-SARI values for the four simplified articles. The fourth simplified article does the best job of simplification, not only retaining the main meaning of the original article but also deleting the unimportant information and the difficult words. However, its SARI value is the lowest among the simplified articles.}
\label{example_sari}
\end{table*}


\noindent
\textbf{Original article: }
\textit{marengo is a town in and the county seat of iowa county , iowa , 
            united states . it has served as the county seat since august 1845 , 
            even though it was not incorporated until july 1859 . the population 
            was 2,528 in the 2010 census , a decline from 2,535 in 2000 .
}

\noindent
\textbf{Simplified article 1: }
\textit{in the US . 2,528 in 2010 .}

\noindent
\textbf{Simplified article 2: }
\textit{marengo is a city in iowa , the US . it has served as the county seat since august 1845 , even though it was not incorporated . the population was 2,528 in the 2010 census , a decline from 2,535 in 2010 .}

\noindent
\textbf{Simplified article 3: }
\textit{marengo is a town in iowa . marengo is a town in the US . in the US . the population was 2,528 . the population in the 2010 census .}

\noindent
\textbf{Simplified article 4: }
\textit{marengo is a town in iowa , united states . in 2010 , the population was 2,528 .}

\noindent
\textbf{Reference article: }
\textit{marengo is a city in iowa in the US . the population was 2,528 in 2010 .}

The SARI values of the four simplified articles are shown in the left half of Table \ref{example_sari}\footnote{We use the script in \url{https://github.com/cocoxu/simplification/blob/master/SARI.py}.}. When using the SARI metric, we take the whole article as input, output and reference. 


The simplified article 1 generates ``US", a word that does not appear in the original article, which raises the overall SARI value. Intuitively, however, it makes no sense to generate several simplified words if the simplified article is too short to convey the main meaning of the original article. Therefore, we believe that a penalty factor $\text{LP}_{1}$ should be added to the $F_{add}$ score. If the generated simplified article is shorter than the reference article, the $F_{add}$ score will be penalized.

The simplified article 2 does a good job of simplifying the first sentence of the original article but retains much useless information and difficult words. Paradoxically, its $P_{del}$ score is the highest among the four simplified essays. A more common scenario is that the original article is much longer than the simplified article, then according to the formula of $P_{del}$, removing fewer words will have a limited effect on $P_{del}$. Therefore, we believe that a penalty factor $\text{LP}_2$ should be added to the $P_{del}$ score, penalizing the $P_{del}$ score if the generated simplified article is longer than the referenced article.

The simplified article 3 finds the important information in the original article and does a good job of simplifying it. Nevertheless, it performs duplicate generation, which leads to a severe decrease in readability and should not get such a high $F_{keep}$ value. According to the formula of $F_{keep}$, if the duplicate n-grams also appear in the reference, then the $F_{keep}$ value will not decrease. Therefore, we believe that the $\text{LP}_2$ should also be added to the $F_{keep}$ score. Besides, we add a sentence-level penalty factor SLP to penalize the $F_{keep}$ score if the number of generated sentences is far from the number of sentences in the reference article.

\begin{table*}[h]
\centering
\resizebox{\textwidth}{!}{
\begin{tabular}{|l|cccc|cccc|c|c|}
\hline
\multicolumn{1}{|c|}{} & D-SARI$\uparrow$         & $D_{keep}$           & $D_{del}$            & $D_{add}$            & SARI$\uparrow$           & $F_{keep}$           & $P_{del}$            & $F_{add}$            & BLEU$\uparrow$           & FKGL$\downarrow$           \\ \hline
Transformer            & {\ul 37.38}    & 31.30          & {\ul 68.80} & {\ul 12.04}    & 44.46    & 43.14          & 75.07 & {\ul 15.16}    & 21.70          & \textbf{17.83} \\
SUC                    & 12.92          & 13.05          & 22.27          & 3.44           & 34.13          & 39.78          & 59.05          & 3.56           & 18.13          & 59.31          \\
BertSumextabs          & \textbf{39.88} & \textbf{35.71} & \textbf{72.06}    & 11.87          & {\ul 47.39}    & \textbf{50.68} & {\ul 76.98}          & 14.50          & {\ul 26.96}    & {\ul 18.32}    \\
BART                   & 37.24          & {\ul 34.34}    & 62.41          & \textbf{14.98} & \textbf{48.34} & {\ul 50.50}    & \textbf{77.72}    & \textbf{16.80} & \textbf{31.77} & 31.90          \\ \hline
\end{tabular}}
\caption{The automatic evaluation results on the D-Wikipedia test set. We use \textbf{Bold} to mark the best result and \underline{underline} the second best result.}
\label{automatic-results}
\end{table*}

\begin{table}[h]
\centering
\resizebox{7.5cm}{!}{
\begin{tabular}{|l|cccc|}
\hline
\multicolumn{1}{|c|}{} & D-SARI$\uparrow$ & SARI$\uparrow$   & BLEU$\uparrow$      & FKGL$\downarrow$   \\ \hline
Transformer            & 27.03  & 28.46  & 0.11      & 27.06  \\
SUC                    & 5.80   & 38.37  & 16.40     & 343.42 \\
BertSumextabs          & 27.68  & 30.23  & 0.20      & 26.40  \\
BART                   & 28.56  & 32.29  & 0.64  & 43.39       \\ \hline
\end{tabular}}
\caption{The automatic evaluation results on the Newsela Simp-4 test set. The BART model performs the best in terms of D-SARI, but the results of all models decline to some degree compared to Table \ref{automatic-results}.}
\label{automatic-newsela-results}
\end{table}

\begin{table*}[h]
\centering
\resizebox{14cm}{!}{
\begin{tabular}{|l|ccccc|c|}
\hline
              & \begin{tabular}[c]{@{}c@{}}Simplicity\\ -phrase\end{tabular} & \begin{tabular}[c]{@{}c@{}}Simplicity\\ -structure\end{tabular} & Meaning       & Grammar       & O-simplicity       & \begin{tabular}[c]{@{}c@{}}Average\\ length\end{tabular} \\ \hline
Transformer   & \textbf{4.77}                                                & \textbf{4.74}                                                   & 2.76          & 4.50          & 2.79          & 58.28                                                    \\
SUC           & 3.09                                                         & 2.95                                                            & \textbf{4.57} & 3.78          & 2.63          & 194.39                                                   \\
BertSumextabs & {\ul 4.61}                                                   & {\ul 4.50}                                                      & 3.60          & \textbf{4.70} & {\ul 3.62}    & 44.21                                                    \\
BART          & 4.06                                                         & 4.00                                                            & {\ul 4.23}    & \textbf{4.70} & 3.59          & 86.42                                                    \\ \hline
Reference     & 4.28                                                         & 4.23                                                            & 3.69          & {\ul 4.65}    & \textbf{3.94} & 81.46                                                    \\ \hline
\end{tabular}}
\caption{The results of human evaluation on the 100 selected article pairs. We use \textbf{Bold} to mark the best result and \underline{underline} the second-best result. The five-point Likert scale is used for rating.}
\label{tab:human_evaluation}
\end{table*}

In summary, based on the SARI metric \cite{xu2016optimizing}, we propose the D-SARI metric for the document-level simplification task. We retain the idea of calculating the scores of add, keep and delete separately in SARI, which proved to be effective in sentence simplification. The D-SARI metric is shown as below:

\begin{equation}
\begin{aligned}
     \text{LP}_{1}&=\left\{
\begin{array}{lcl}
1       &      & O \geq R\\
e^{\frac{O-R}{O}}     &      & O < R
\end{array} \right.   \\
    \text{LP}_{2}&=\left\{
\begin{array}{lcl}
1       &      & O \leq R\\
e^{\frac{R-O}{max(I-R,1)}}     &      & O > R
\end{array} \right.
\end{aligned}
\end{equation}
\begin{equation}
    \text{SLP} \ \ \ =e^{-\frac{||R_S-O_S||}{max(R_S,O_S)}}
\end{equation}
\begin{equation}
\begin{aligned}
    & D_{keep} = F_{keep} * \text{LP}_{2} * \text{SLP} \\
    & D_{add} \ = F_{add} * \text{LP}_{1} \\
    & D_{del} \ \ = P_{del} * \text{LP}_{2}
\end{aligned}
\end{equation}
\begin{equation}
\begin{aligned}
    \text{D-SARI} = \ & ( D_{keep} + D_{del} + D_{add} ) * 1/3
\end{aligned}
\end{equation}

$I$, $O$, and $R$ represent the number of words (including punctuation) in the input article, the output article, and the reference article, respectively. $O_{S}$ and $R_{S}$ represent the number of sentences in the output article and the reference article, respectively. Due to the limitation of space, please refer to \newcite{xu2016optimizing} for the calculation of $F_{add}$, $F_{keep}$ and $P_{del}$. We also calculate the D-SARI values for each of the simplified articles in the given example, as shown in the right half of Table \ref{example_sari}. 

As we analyzed from the given example, it is reasonable to penalize the three components in SARI. In the D-SARI metric, the penalty is based on length. The motivation comes from BLEU \cite{papineni2002bleu}. A candidate should be neither too long nor too short, and an evaluation metric should enforce this. The difference between the length of a simplified sentence and the original sentence in sentence-level text simplification is not very large, while the opposite is true for document-level text simplification. An original article may be long, while a simplified article may contain only one sentence. It is simple enough, but not a good simplification of the original article. It is a reasonable proposition that the length of the simplified article should be close to the length of the reference article.

We also conduct an empirical analysis of the D-SARI metric. In Section \ref{sec:correlation}, we use Spearman's rank correlation coefficient \cite{zwillinger1999crc} to show that the D-SARI metric has the strongest correlation among several metrics with human ratings.








\begin{table*}[h]
\centering
\begin{tabular}{|l|ccccc|}
\hline
Spearman's $\rho$ & \begin{tabular}[c]{@{}c@{}}Simplicity\\ -phrase\end{tabular} & \begin{tabular}[c]{@{}c@{}}Simplicity\\ -structure\end{tabular} & Meaning       & Grammar       & O-simplicity       \\ \hline
BLEU       & -0.14                                                        & -0.12                                                           & \textbf{0.23} & 0.09          & 0.30          \\
SARI       & 0.28                                                         & 0.34                                                            & -0.22         & 0.29          & 0.36          \\
-FKGL      & \textbf{0.66}                                                & \textbf{0.67}                                                   & -0.63         & \textbf{0.47} & 0.18          \\
D-SARI     & 0.42(+0.14)                                                         & 0.47(+0.13)                                                            & -0.30(-0.08)         & 0.38(+0.09)          & \textbf{0.42}(+0.06) \\ \hline
\end{tabular}
\caption{Correlation of the automatic metrics against the human ratings. We use \textbf{Bold} to mark the best result. Because the simpler the article, the lower the FKGL value, we report -FKGL for better comparison. The differences between SARI and D-SARI are also shown in brackets.}
\label{tab:correlation_metrics}
\end{table*}

\section{Baseline Models}


We selected four representative models as the baselines for the document-level simplification task, which are:


(1) Transformer: It treats the task as a sequence-to-sequence problem. Both the encoder and decoder contain six transformer layers \cite{vaswani2017attention}. 

(2) SUC: It simplifies each sentence in the article by using use contextual information \cite{sun2020helpfulness}.

(3) BertSumextabs: It achieves excellent results on the text summarization task, using the Bert-base model as the encoder \cite{liu2019text}.


(4) BART: It is a recently proposed pre-trained model based on large-scale corpus and achieves state-of-the-art results on many sequence-to-sequence tasks \cite{lewis2019bart}. 

All the models were tested on our delineated test sets. We used the fairseq toolkit and performed replicate experiments. See Appendix \ref{section:appendixB} for detailed parameters.


\section{Evaluation Results}

\subsection{Automatic Evaluation Results}

We used the SARI metric, the BLEU metric, the FKGL metric, and the D-SARI metric for automatic evaluation. We have described the SARI and D-SARI metrics in detail in Section \ref{sec:criteria}. BLEU is a method for comparing the similarity between the reference and the output \cite{papineni2002bleu}\footnote{We used the script in \url{https://github.com/nltk/nltk/blob/develop/nltk/translate/bleu_score.py}.}. FKGL is used to measure the readability of the text \cite{kincaid1975derivation}\footnote{We used the script in \url{https://github.com/shivam5992/textstat/blob/master/textstat/textstat.py}.}. 

The automatic evaluation results of the D-Wikipedia test set are shown in Table \ref{automatic-results}. The BertSumextabs model obtains the best results on the D-SARI value. The BART model obtains the best results on the SARI value and the BLEU value. The transformer model obtains the best results on the FKGL value. We also give an example to compare the outputs of different models, and we put it in Appendix \ref{section:appendixD}.

\begin{table}[h]
\centering
\begin{tabular}{|l|c|}
\hline
Spearman's $\rho$          & \begin{tabular}[c]{@{}c@{}}The length of \\ simplified article\end{tabular} \\ \hline
Simplicity-phrase    & -0.65                                                                       \\
Simplicity-structure & -0.65                                                                       \\
Meaning              & 0.56                                                                        \\
Grammar              & -0.50                                                                       \\
O-simplicity              & -0.26                                                                       \\ \hline
\end{tabular}
\caption{Correlation of the simplified article's length against the human ratings. The simplicity-phrase and simplicity-structure score have a strong negative correlation with the article's length, while the O-simplicity score has a weak correlation.}
\label{tab:correlation_length}
\end{table}

For the Newsela corpus, as mentioned in Section \ref{sec:Newsela}, we choose a representative test set called Simp-4 to show the automatic results. The models were both trained and validated on the D-Wikipedia dataset, and the results are shown in Table \ref{automatic-newsela-results}.

\subsection{Human Evaluation Results}
\label{sec:human_evaluation}

We performed human evaluation according to the method described in Section \ref{sec:dataset_human}. To maintain consistency, we selected the same 100 article pairs in the D-Wikipedia test set that were randomly selected for evaluating the dataset in Section~\ref{sec:dataset_human}. We added some fake examples to the questionnaire and checked whether the workers gave a reasonable score to ensure the quality of human evaluation.

The human evaluation results are shown in Table \ref{tab:human_evaluation}. We also report the correlation of the article's length against the human ratings, as shown in Table \ref{tab:correlation_length}. The results prove that human judges tend to give high simplicity-phrase scores and high simplicity-structure scores to short articles.
The O-simplicity indicator places more emphasis on the overall simplification effect, including the retention of the main meaning and the fluency of the sentences.
Therefore, as we analyzed in Section \ref{sec:dataset_human}, the O-simplicity indicator can evaluate how good the simplification is, which is better than the simplicity-phrase and the simplicity-structure indicators. Generally, the BART and BertSumextabs models perform better than the other two models, especially on the O-simplicity measure. Directly applying the sentence simplification model SUC does not get good results, which means document-level simplification is very different from sentence-level simplification. 



\subsection{Correlation of Automatic Metrics with Human Ratings}
\label{sec:correlation}

We calculated Spearman's rank correlation coefficient between each automatic metric and human ratings on the results for the 100 article pairs, and the correlation scores are shown in Table \ref{tab:correlation_metrics}.
The D-SARI metric has the highest correlation with the O-simplicity indicator, surpassing both BLEU and SARI. In terms of simplicity-phrase and simplicity-structure, the correlation of D-SARI with human ratings also exceeds that of SARI, and although FKGL has the highest correlation, it does not correlate with the O-simplicity indicator. We also noticed that BLEU has little correlation with the meaning and grammar indicators, probably because the simplification contains lots of splitting operations, which is consistent with the conclusion obtained by \newcite{sulem2018bleu}.

\subsection{The Challenge of Document-level Simplification}

There are many problems with applying existing models directly to the document-level simplification task. From the automatic evaluation, the $D_{keep}$ values of the baseline models are not high, and the FKGL values also need to be further reduced. From human evaluation, the O-simplicity scores of the articles simplified by the models are still far from that of the reference. 

As can be seen from the given example in Appendix \ref{section:appendixD}, the best-performing BertSumextabs model among the four models still retains some complex vocabulary and sentence structure compared with the reference, and the model's ability to screen out important information needs further improvement. We also noticed that the results of the SUC model are much lower than all other models, which indicates that document-level simplification cannot be addressed by stitching together the results of sentence simplification as simplified articles.

Above all, we believe that new models designed for document-level simplification could be proposed in the future, which will greatly advance this field.

\section{Conclusion}


In this paper, we are committed to promoting research on document-level text simplification. We established a large-scale high-quality dataset named D-Wikipedia and proposed a new automatic evaluation metric called D-SARI. We also selected several representative models as baselines for this task. The results demonstrate that the dataset is of high quality and the metric is reliable.

\section*{Acknowledgements}

We are grateful to the reviewers for their valuable comments.

This work was supported by National Natural Science Foundation of China (61772036), Beijing Academy of Artificial Intelligence (BAAI) and Key Laboratory of Science, Technology and Standard in Press Industry (Key Laboratory of Intelligent Press Media Technology). Xiaojun Wan is the corresponding author.




\bibliography{anthology,custom}
\bibliographystyle{acl_natbib}

\appendix


\section{Six Types of Operations in Document-Level Text Simplification}
\label{section:appendixA}

\textbf{1 Sentence joining and sentence reordering}

\noindent
\textbf{Src:} the fields medal is a prize awarded to two , three , or four mathematicians under 40 years of age at the international congress of the international mathematical union ( imu ) , a meeting that takes place every four years. \textcolor{red}{the fields medal is regarded as one of the highest honors a mathematician can receive} , and has been described as the mathematician 's nobel prize , although there are several key differences , including frequency of award , number of awards , and age limits . \textcolor{red}{according to the annual academic excellence survey by arwu , the fields medal is consistently regarded as the top award in the field of mathematics worldwide} , and in another reputation survey conducted by ireg in 2013–14 , the fields medal came closely after the abel prize as the second most prestigious international award in mathematics. the prize comes with a monetary award which , since 2006 , has been 15,000 . the name of the award is in honour of canadian mathematician john charles fields . fields was instrumental in establishing the award , designing the medal itself , and funding the monetary component. the medal was first awarded in 1936 to finnish mathematician lars ahlfors and american mathematician jesse douglas , and it has been awarded every four years since 1950 . \underline{its purpose is to give recognition and support to y-}
\underline{ounger mathematical researchers who have made }
\underline{major contributions .}

\noindent
\textbf{Tgt:} the fields medal is a prize given to mathematicians who are not over 40 years of age . it is given at each international congress of the international mathematical union . this is a meeting that takes place every four years. the canadian mathematician john charles fields was the first to propose this medal and it was first awarded in 1936 . it has been regularly awarded since 1950 . \underline{its purpose is to support younger mathematicians }
\underline{who made major contributions.} \textcolor{red}{the fields medal is viewed , at least in the media , as the top honor a mathematician can receive .}

\noindent
\textbf{Analysis:} Sentence joining means combining two or more sentences into one sentence. In src, the two sentences marked in red are merged into the sentence marked in red in tgt. Some information in the original two sentences is removed. Sentence reordering implies a change in the structure of the article. In src, the sentences marked in red appear before the sentence marked as underlined, but in tgt, the simplified sentence marked as underlined appears before the sentence marked in red.

~\\
\noindent
\textbf{2 Sentence splitting}

\noindent
\textbf{Src:} \textcolor{red}{it is a decentralized digital currency} \textcolor{blue}{without a central bank or single administrator} \textcolor{cyan}{that can be sent from user to user on the peer-to-peer bitcoin network} \textcolor{blue}{without the need for intermediaries .}

\noindent
\textbf{Tgt:} \textcolor{red}{bitcoin is a digital and global money system currency .} \textcolor{cyan}{it allows people to send or receive money across the internet , even to someone they do n't know or do n't trust .} \textcolor{blue}{money can be exchanged without being linked to a real identity .}

\noindent
\textbf{Analysis:} In contrast to sentence joining, sentence splitting is the division of a long sentence into two or more sentences. The sentences in tgt are simplified from the parts of the sentence in src marked with the corresponding colors.

~\\
\noindent
\textbf{3 Sentence addition}

\noindent
\textbf{Src:} 104.6 rtl is a private radio station that is produced in a hot adult contemporary format . it is transmitted from studios in kurfürstendamm in berlin-charlottenburg . according to german media analysis 2011/ii , the station reaches 209,000 listeners in an average transmitting hour ( mon-fri , 6am-6pm ) with a total of 709,000 listeners per day and thereby is one of the most listened to radio programs in berlin and brandenburg .

\noindent
\textbf{Tgt:} 104.6 rtl is a german radio station . \textcolor{red}{it first aired on 9 september , 1991 .} it broadcasts in berlin and hopes that the 14-39 age group will listen . the studios are at the kurfürstendamm in berlin-charlottenburg . \textcolor{red}{in 2005 the radio channel has been awarded the german radio award for the best morning show .}

\noindent
\textbf{Analysis:} Sentence addition means that there is a sentence in tgt for which the corresponding sentence is not found in src. Sentence addition introduces additional information, often used for explanation and clarification.

~\\
\noindent
\textbf{4 Sentence deletion}

\noindent
\textbf{Src:} landudal is a commune in the finistère department of brittany in north-western france . \textcolor{red}{the writer angèle jacq , winner of the cezam prix littéraire inter ce in 2000 for her novel `` le voyage de jabel '' , was born in landudal .}

\noindent
\textbf{Tgt:} landudal is a commune . it is found in the region brittany in the finistère department in the northwest of france .

\noindent
\textbf{Analysis:} Sentence deletion means that there is a sentence in src that does not find a corresponding sentence in tgt. This is usually because the original sentence is difficult to simplify and the deletion does not affect the main meaning of the text.

~\\
\noindent
\textbf{5 Anaphora resolution}

\noindent
\textbf{Src:} \textcolor{red}{Winston Churchill} was a great politician and statesman. \textcolor{red}{He} also won the Nobel Prize for literature in 1953.

\noindent
\textbf{Tgt:} \textcolor{red}{Winston Churchill} won the Nobel Prize in 1953.

\noindent
\textbf{Analysis:} Anaphora resolution is usually associated with sentence deletion. the first sentence in src is deleted, then the pronoun “he” in the second sentence is replaced with the person's name in tgt.

\section{Implementation Details}
\label{section:appendixB}

We used the fairseq toolkit\footnote{\url{https://github.com/pytorch/fairseq}} to implement the transformer model and the BART model. We used the code on the github to implement the BertSumextabs model\footnote{\url{https://github.com/nlpyang/PreSumm}} and the SUC model\footnote{\url{https://github.com/RLSNLP/Document-Context-to-Sentence-Simplification}}. All the models except the SUC model are trained on the training set of the D-Wikipedia dataset we constructed. The SUC model is trained on the Wikipedia dataset and when testing, the original articles in our delineated test set are simplified with this model sentence by sentence, and then the output sentences are stitched together to get the simplified articles. All the models are trained on Nvidia GTX 1080ti. The batchsize we set can make full use of its video memory. The hyperparameters are shown in the following tables.

\begin{table}[h]
\centering
\begin{tabular}{|l|c|}
\hline
hyperparameter  & value \\ \hline
learning rate   & 1e-3  \\
dropout         & 0.1   \\
max tokens      & 2048  \\
update freq     & 4     \\
label smoothing & 0.1   \\
weight decay    & 1e-4  \\
num updates     & 1e5   \\ \hline
\end{tabular}
\caption{The hyperparameters of the transformer model.}
\end{table}

\begin{table}[h]
\centering
\begin{tabular}{|l|c|}
\hline
hyperparameter  & value \\ \hline
learning rate   & 1e-4  \\
dropout         & 0.1   \\
max tokens      & 2048  \\
update freq     & 4     \\
label smoothing & 0.1   \\
weight decay    & 1e-4  \\
num updates     & 1e5   \\ \hline
\end{tabular}
\caption{The hyperparameters of the BART model.}
\end{table}

\begin{table}[h]
\centering
\begin{tabular}{|l|c|}
\hline
hyperparameter      & value \\ \hline
learning rate       & 0.1   \\
activation function & GELU  \\
batchsize           & 16    \\
encoder layers      & 4     \\
decoder layers      & 4     \\
multi-heads         & 4     \\
max epochs          & 50    \\ \hline
\end{tabular}
\caption{The hyperparameters of the SUC model.}
\end{table}

\begin{table}[h]
\centering
\begin{tabular}{|l|c|}
\hline
hyperparameter    & value \\ \hline
max learning rate & 2e-3  \\
dropout           & 0.1   \\
batchsize         & 500    \\
update fraq       & 8    \\
num updates       & 50000 \\ \hline
\end{tabular}
\caption{The hyperparameters of the BertSumextabs model (ext).}
\end{table}

\begin{table}[h!]
\centering
\begin{tabular}{|l|c|}
\hline
hyperparameter             & value \\ \hline
max learning rate(encoder) & 2e-3  \\
max learning rate(decoder) & 0.2   \\
dropout                    & 0.2   \\
batchsize                  & 12    \\
update fraq                & 20    \\
num updates                & 50000 \\ \hline
\end{tabular}
\caption{The hyperparameters of the BertSumextabs model (abs).}
\end{table}

\section{Human Evaluation Guideline}
\label{section:appendixC}


The goal of this review is to evaluate the simplification quality of different articles. In this review, you will be given an original article and its corresponding simplified articles. You should evaluate the quality of the simplification in the following five ways:

\noindent
(1)	\textbf{Simplicity-phrase.} Are the words in the simplified article simpler than those in the original article?

\noindent
(2)	\textbf{Simplicity-structure.} Are the sentence structures in the simplified article simpler than those in the original article?

\noindent
(3)	\textbf{Meaning.} The text simplification operation can remove some sentences from the original article, but the main meaning of the original article should be kept intact.

\noindent
(4) \textbf{Grammar.} The simplified article should be grammatically correct and fluent.

\noindent
(5) \textbf{O-simplicity.} The simplified article should be simpler than the original article, and it also should read smoothly and can retain the main meaning of the original article.

\noindent
You will do this using a 1-5 rating scale, where 5 is the best and 1 is the worst. There are no “correct” answers and whatever choice is appropriate for you is a valid response. For example, if you are given the following original article and simplified articles:
~\\

\noindent
\textbf{Original article:}
the fields medal is a prize awarded to two, three, or four mathematicians under 40 years of age at the international congress of the international mathematical union ( imu ), a meeting that takes place every four years. the fields medal is regarded as one of the highest honors a mathematician can receive, and has been described as the mathematician 's nobel prize, although there are several key differences, including frequency of award, number of awards, and age limits. according to the annual academic excellence survey by arwu, the fields medal is consistently regarded as the top award in the field of mathematics worldwide, and in another reputation survey conducted by ireg in 2013–14, the fields medal came closely after the abel prize as the second most prestigious international award in mathematics. the prize comes with a monetary award which, since 2006, has been 15,000. the name of the award is in honour of canadian mathematician john charles fields. fields was instrumental in establishing the award, designing the medal itself, and funding the monetary component. the medal was first awarded in 1936 to finnish mathematician lars ahlfors and american mathematician jesse douglas, and it has been awarded every four years since 1950. its purpose is to give recognition and support to younger mathematical researchers who have made major contributions. in 2014, the iranian mathematician maryam mirzakhani became the first female fields medalist. in all, sixty people have been awarded the fields medal. the most recent group of fields medalists received their awards on 1 august 2018 at the opening ceremony of the imu international congress, held in rio de janeiro, brazil. the medal belonging to one of the four joint winners, caucher birkar, was stolen shortly after the event. the icm presented birkar with a replacement medal a few days later.

\noindent
\textbf{Simplified article 1: \ }
\textbf{(Score: Simplicity-phrase 5 Simplicity-structure 5 Meaning 5 Grammar 5 O-simplicity 5)}

\noindent
the fields medal is an award given to mathematicians under 40 years of age. the name of the prize is in honor of the canadian mathematician john charles field. and it is awarded every four years since 1950. the fields medal is regarded as the highest award in the field of mathematics in the world.  it is intended to be used to encourage young mathematicians.

\noindent
\textbf{Simplified article 2: \ }
\textbf{(Score: Simplicity-phrase 4 Simplicity-structure 5 Meaning 5 Grammar 5 O-simplicity 5)}

\noindent
the fields medal is a prize given to mathematicians who are not over 40 years of age. it is given at each international congress of the international mathematical union. this is a meeting that takes place every four years. the canadian mathematician john charles fields was the first to propose this medal and it was first awarded in 1936. it has been regularly awarded since 1950. its purpose is to support younger mathematicians who made major contributions. the fields medal is viewed, at least in the media, as the top honor a mathematician can receive. it comes with a monetary award. in 2006 the award was \$ 15,000 ( us \$ 13,400 or €10,550 ). the abel prize has similar prestige, and more money.

\noindent
\textbf{Simplified article 3: \ }
\textbf{(Score: Simplicity-phrase 4 Simplicity-structure 3 Meaning 2 Grammar 5 O-simplicity 2)}

\noindent
the fields medal is consistently regarded as the top award in the field of mathematics worldwide.  Since 2006, the prize of this award has been 15,000. the most recent group of fields medalists received their awards on 1 august 2018 at the opening ceremony of the imu international congress. the medal belonging to one of the four joint winners, caucher birkar , was stolen shortly after the event .

\noindent
\textbf{Simplified article 4: \ }
\textbf{(Score: Simplicity-phrase 1 Simplicity-structure 1 Meaning 5 Grammar 5 O-simplicity 1)}

\noindent
the fields medal is a prize awarded to two, three, or four mathematicians under 40 years of age at the international congress of the international mathematical union ( imu )  a meeting that takes place every four years. according to the annual academic excellence survey by arwu, the fields medal is consistently regarded as the top award in the field of mathematics worldwide, and in another reputation survey conducted by ireg in 2013–14, the fields medal came closely after the abel prize as the second most prestigious international award in mathematics. the name of the award is in honour of canadian mathematician john charles fields. he was instrumental in establishing the award, designing the medal itself, and funding the monetary component. the purpose of the fields medal is to give recognition and support to younger mathematical researchers who have made major contributions.

\noindent
\textbf{Simplified article 5: \ }
\textbf{(Score: Simplicity-phrase 5 Simplicity-structure 5 Meaning 4 Grammar 1 O-simplicity 2)}

\noindent
the fields medal is to a prize giving to mathematicians who are not over 40 years of age. but it is awarding every four years since 1950. the prize is in honor of the canadian mathematician john charles field. the fields metal described as the mathematician 's nobel prize as the mathematician 's nobel prize. its purpose are to support younger mathematicians who made major contributions.

~\\
\noindent
\textbf{Analysis:}
The \textbf{Simplified article 1} does a good job on the simplification of words and sentence structures. The simplification includes removing difficult vocabulary, splitting and simplifying long sentences, etc.. So, it scores full marks for the simplification-phrase and the simplification-structure. It is equally able to summarize the main meaning of the original article, so it scores full marks for meaning. It reads smoothly, like it is written by humans, so it scores full marks for grammar. The overall feeling of the article is very good. It reads very simple, fluently and maintains the main meaning, so it scores full marks for the O-simplicity.
The \textbf{Simplified article 2} has some words that need further simplification, such as “prestige” and “monetary”. So, it scores a little bit lower than the simplified article 1 on the simplicity-phrase. However, it reads smoothly and the main meaning is well maintained. One will also feel that the simplification effect is very good when reading this article. These two articles also illustrate that articles that score high marks can be presented in different ways.
Obviously, \textbf{simplified article 3} does not retain the main meaning of the original article, but rather some non-essential information. Therefore, it scores very low on meaning. Besides, it contains long and complex sentences and the sentence structures are not simple enough compared to the original article. One' s experience of reading such an article is not very good, because it deviates from the main meaning and is not simple enough.
The \textbf{Simplified article 4} is able to find those relatively important sentences in the original article. But unfortunately, it does little simplification operation and is not easy to read, so it scores very low on the simplification-phrase and the simplification-structure. Children and non-native speakers will not be able to read such an article, so it scores very low on the O-simplicity.
The \textbf{Simplified article 5} contains many grammatical errors and repetition of some phrases, making it look less like it is written by a human. Therefore, it scores very low on grammar. Although its words and sentence structures are very simple, the existence of grammatical errors makes it difficult to read, so it scores low on the O-simplicity.



\section{Case Study}
\label{section:appendixD}



\textbf{Input:} atal bihari vajpayee ( ; 25 december 1924 – 16 august 2018 ) was an indian statesman who served three terms as the prime minister of india , first for a term of 13 days in 1996 , then for a period of 13 months from 1998 to 1999 , followed by a full term from 1999 to 2004 . a member of the bharatiya janata party ( bjp ) , he was the first indian prime minister not of the indian national congress to serve a full term in office . he was also noted as a poet and a writer . he was a member of the indian parliament for over five decades , having been elected ten times to the lok sabha , the lower house , and twice to the rajya sabha , the upper house . he served as the member of parliament for lucknow , retiring from active politics in 2009 due to health concerns . he was among the founding members of the bharatiya jana sangh ( bjs ) , of which he was president from 1968 to 1972 . the bjs merged with several other parties to form the janata party , which won the 1977 general election . in march 1977 , vajpayee became the minister of external affairs in the cabinet of prime minister morarji desai . he resigned in 1979 , and the janata alliance collapsed soon after . former members of the bjs formed the bjp in 1980 , with vajpayee its first president . during his tenure as prime minister , india carried out the pokhran - ii nuclear tests in 1998 . vajpayee sought to improve diplomatic relations with pakistan , travelling to lahore by bus to meet with prime minister nawaz sharif . after the 1999 kargil war with pakistan , he sought to restore relations through engagement with president pervez musharraf , inviting him to india for a summit at agra . the administration of narendra modi declared in 2014 that vajpayee 's birthday , 25 december , would be marked as good governance day . in 2015 , he was conferred india 's highest civilian honour , the bharat ratna , by the president of india , pranab mukherjee . he died on 16 august 2018 of age - related illness .

~\\
\noindent
\textbf{Reference:}  atal bihari vajpayee ( 25 december 1924 – 16 august 2018 ) was an indian politician . he served as the 10th prime minister of india for six years from 1998 through 2004 . he was also serving as prime minister from may 1996 through june 1996 . he founded the bharatiya janata party . vajpayee was born on 25 december 1924 in gwalior , united provinces , british india ( now madhya pradesh , india ) . he studied at victoria college ( now laxmibai college ) and at dav college , kanpur . vajpayee was never married and had no children . vajpayee , who had dementia and diabetes , died at a new delhi hospital from multiple organ failure on 16 august 2018 , aged 93 .

~\\
\noindent
\textbf{Transformer:}  \textcolor{red}{amar prasad} ( 25 december 1924 – 16 august 2018 ) was an indian politician . he was a member of the indian national congress party . he served as prime minister of india from 1999 to 2004 . \textcolor{red}{he was also the first prime minister of india .} he was born in <unk> , <unk> died on 16 august 2018 at the age of 93 .

~\\
\noindent
\textbf{SUC:}  atal bihari vajpayee ( ; 25 december 1924 – 16 august 2018 ) was an indian statesman who served three terms as the prime minister of india , first for a term of 13 days in 1996 , then for a period of 13 months from 1998 to 1999 , followed by a full term from 1999 to 2004 . a member of the bharatiya janata party ( bjp ) , \textcolor{red}{he was the first indian prime minister not of the indian national congress to be elected into the full term in the first indian prime minister .} he was the first indian prime minister of the indian national congress and was called the bharatiya janata party ( bjp ) . he was a member of the indian parliament for over five decades and has been elected ten times to the lok sabha , the lower house , and twice to the rajya sabha , the upper house . he was elected ten times to the lok sabha , the lower house , and twice to the rajya sabha , the upper house . he served as the member of parliament for lucknow , retiring from active politics in 2009 due to health concerns . \textcolor{blue}{he was among the founding members of the bharatiya jana sangh ( bjs ) , of which he was president from 1968 to 1972 . the bjs merged with several other parties to form the janata party , which won the 1977 general election .} \textcolor{cyan}{it to make the janata party , he won the 1977 general election .} \textcolor{blue}{in march 1977 , vajpayee became the minister of external affairs in the cabinet of prime minister morarji desai .} he became the minister of external affairs in the cabinet of prime minister morarji desai . he left in 1979 and left the alliance collapsed soon after . \textcolor{cyan}{in 1980 's first president.}during members of prime minister , india carried out the bjp nuclear tests as prime minister . in 1998 he became prime minister , india , and created the pokhran-ii nuclear tests . \textcolor{blue}{vajpayee sought to improve diplomatic relations with pakistan , travelling to lahore by bus to meet with prime minister nawaz sharif . after the 1999 kargil war with pakistan , he sought to restore relations through engagement with president pervez musharraf , inviting him to india for a summit at agra.the administration of narendra modi declared in 2014 that vajpayee 's birthday , 25 december , would be marked as good governance day .} in 2015 , he was conferred india 's highest civilian honour , the bharat ratna , by the president of india , pranab mukherjee . he died on 16 august 2018 of age-related illness .

~\\
\noindent
\textbf{BertSumextabs:}  atal bihari vajpayee ( ; 25 december 1924 – 16 august 2018 ) was an indian politician . he served three terms as the prime minister of india , first for a term of 13 days in 1996 , then for a period of 13 months from 1998 to 1999 , followed by a full term from 1999 to 2004 . \textcolor{cyan}{a member of the bharatiya janata party ( bjp ) ,} \textcolor{red}{he was the first indian prime minister not of the indian national congress .}

~\\
\noindent
\textbf{BART:}  atal bihari vajpayee ( 25 december 1924 – 16 august 2018 ) was an indian politician . he was the first indian prime minister not of the indian national congress to serve a full term in office.\textcolor{blue}{he was elected as a member of the indian parliament for over five decades , having been elected ten times to the lok sabha , the lower house , and twice to the rajya sabha , the upper house . he served as the member of parliament for lucknow , retiring from active politics in 2009 due to health concerns.he was among the founding members of the bharatiya jana sangh ( bjs ) , of which he was president from 1968 to 1972.in march 1977 , vajpayee became the minister of external affairs in the cabinet of prime minister morarji desai . he resigned in 1979 , and the janata alliance collapsed soon after .} former members of the bjs formed the bjp in 1980 , with vajpayee its first president.during his term as prime minister , india carried out the pokhran-ii nuclear tests in 1998 . vajpayee sought to improve diplomatic relations with pakistan , travelling to lahore by bus to meet prime minister nawaz sharif . \textcolor{blue}{after the 1999 kargil war with pakistan , he sought to restore relations through engagement with president pervez musharraf , inviting him to india for a summit at agra.} vajpayee died on 16 august 2018 \textcolor{red}{in lucknow }, aged 93 .



~\\
\noindent
\textbf{Analysis:} We use \textcolor{red}{red} to mark sentences with factual errors. We use \textcolor{blue}{blue} to mark sentences that should have been deleted but are not deleted and not simplified, and we use \textcolor{cyan}{cyan} to mark sentences with grammatical errors. The output articles of the SUC model and the BART model are too long and retain a large number of unsimplified sentences in the 
input article. The output article of the Transformer model contains many factual errors and is poorly readable. The BertSumextabs model simplifies a less important sentence in the original article, and the simplification is not reasonable. Because it removes the critical information of ``to serve a full term in office", the meaning of the sentence may be changed. Besides, 
the BertSumextabs model do not keep the information about the person's death from the original article.

\end{document}